\documentclass[sigconf, nonacm]{acmart}

\usepackage{subcaption}

\copyrightyear{2021} 
\acmYear{2021} 
\setcopyright{acmlicensed}\acmConference[ICMI '21 Companion]{Companion Publication of the 2021 International Conference on Multimodal Interaction}{October 18--22, 2021}{Montréal, QC, Canada}
\acmBooktitle{Companion Publication of the 2021 International Conference on Multimodal Interaction (ICMI '21 Companion), October 18--22, 2021, Montréal, QC, Canada}
\acmPrice{15.00}
\acmDOI{10.1145/3461615.3485423}
\acmISBN{978-1-4503-8471-1/21/10}

\begin{document}


\title[BERT meets LIWC]{BERT meets LIWC: Exploring State-of-the-Art Language Models for Predicting Communication Behavior in Couples’ Conflict Interactions}






\author{Jacopo Biggiogera}
\affiliation{%
  \institution{University of Surrey}
  \city{Guildford}
  \country{U.K.}}
\email{jb01493@surrey.ac.uk}

\author{George Boateng}
\affiliation{%
  \institution{ETH Zurich}
  \city{Zurich}
  \country{Switzerland}}
\email{gboateng@ethz.ch}

\author{Peter Hilpert}
\affiliation{%
  \institution{University of Lausanne}
  \city{Lausanne}
  \country{Switzerland}}
\email{peter.hilpert@unil.ch}

\author{Matthew Vowels}
\affiliation{%
  \institution{University of Surrey}
  \city{Guildford}
  \country{U.K.}}
\email{m.j.vowels@surrey.ac.uk}

\author{Guy Bodenmann}
\affiliation{%
  \institution{University of Zurich}
  \city{Zurich}
  \country{Switzerland}}
\email{guy.bodenmann@psychologie.uzh.ch}

\author{Mona Neysari}
\affiliation{%
  \institution{University of Zurich}
  \city{Zurich}
  \country{Switzerland}}
\email{m.neysari@psychologie.uzh.ch}

\author{Fridtjof Nussbeck}
\affiliation{%
  \institution{University of Konstanz}
  \city{Konstanz}
  \country{Germany}}
\email{fridtjof.nussbeck@uni-konstanz.de}

\author{Tobias Kowatsch}
\email{tkowatsch@ethz.ch}
\affiliation{%
  \institution{ETH Zurich}
  \city{Zurich}
  \country{Switzerland}
}
\affiliation{%
  \institution{University of St. Gallen}
  \city{St. Gallen}
  \country{Switzerland}
}

\renewcommand{\shortauthors}{Biggiogera et al.}

\begin{abstract}
\end{abstract}


\begin{abstract}
Many processes in psychology are complex, such as dyadic interactions between two interacting partners (e.g., patient-therapist, intimate relationship partners). Nevertheless, many basic questions about interactions are difficult to investigate because dyadic processes can be within a person and between partners, they are based on multimodal aspects of behavior and unfold rapidly. Current analyses are mainly based on the behavioral coding method, whereby human coders annotate behavior based on a coding schema. But coding is labor-intensive, expensive, slow, focuses on few modalities, and produces sparse data which has forced the field to use average behaviors across entire interactions, thereby undermining the ability to study processes on a fine-grained scale. Current approaches in psychology use LIWC for analyzing couples’ interactions. However, advances in natural language processing such as BERT could enable the development of systems to potentially automate behavioral coding, which in turn could substantially improve psychological research. In this work, we train machine learning models to automatically predict positive and negative communication behavioral codes of 368 German-speaking Swiss couples during an 8-minute conflict interaction on a fine-grained scale (10-seconds sequences) using linguistic features and paralinguistic features derived with openSMILE. Our results show that both simpler TF-IDF features as well as more complex BERT features performed better than LIWC, and that adding paralinguistic features did not improve the performance. These results suggest it might be time to consider modern alternatives to LIWC, the de facto linguistic features in psychology, for prediction tasks in couples research. This work is a further step towards the automated coding of couples’ behavior which could enhance couple research and therapy, and be utilized for other dyadic interactions as well.
\end{abstract}


\maketitle

\section{Introduction}
There are many processes in the field of psychology that are very complex such as dyadic interactions —  interactions between two people \cite{Hilpert2020}. These processes are difficult to investigate because each person’s behavior is multimodal, both persons influence each other’s behavior mutually, and this process unfolds rapidly \cite{Gottman2005}. Such dynamic processes are relevant for a large number of human interactions (e.g., romantic partners, patient-therapist, student-teacher, buyer-seller). 

Of the different human interactions, conflict interactions in intimate relationships have been well studied over the last decades \cite{Bradbury2010}. Results indicate two principal types of communication behaviors: functional and dysfunctional. For example, contempt and criticism are a reliable predictor for later divorce and therefore seen as negative or dysfunctional, whereas providing appreciation and taking responsibility are considered to be functional and are associated with stable relationships \cite{Gottman1994, Gottman1998, Gottman2018}. It is therefore important to understand conflict interactions better as divorces are not only often emotionally and financially difficult for partners, but also have long-term negative consequences on the children involved \cite{Amato2001}. 

The major reason for the disappointing progress of understanding behavioral processes during conflict interactions is the lack of methods that enable an automated approach for a fine-grained understanding of behavior. Traditionally, analyses in interaction research are mainly undertaken using data obtained from observer rating methods which are labor-intensive, expensive and time-consuming \cite{Kerig2004}. Consequently, codes are generally assigned on a global scale (e.g., one rating for 8-10 minutes sessions) rather than on a fine-grain scale (e.g., every talk turn or 10 seconds) resulting in sparse data. While observer ratings provide a means to capture global aspects of behavior (e.g., positive behavior), the analysis of such global behavioral aspects and sparse data has forced the field to focus on predictions based on average behaviors across entire interactions, thereby undermining the ability to study intra- and inter-individual processes \cite{Hilpert2020}. 

Beyond observer rating methods, psychology has also included technology to extract linguistic (i.e., \textit{what} was said) and paralinguistic features (i.e., \textit{how} it was said). Various paralinguistic features have been extracted mainly using Praat \cite{Boersma2001} and openSMILE \cite{Eyben2010} which are software tools that compute various acoustic features over audio signals (e.g., pitch, fundamental frequency) over sequential time segments (e.g., 25 ms).  They have been used in various works for example to show that the fundamental frequency of oscillation of the vocal folds is a valid proxy for emotional arousal \cite{Juslin2005} and a larger range in fundamental frequency is associated with more conflict interactions \cite{Baucom2011, Baucom2012}. Furthermore, a specific set of 88 features computed called eGeMAPS have been shown to be a minimalist feature set that performs well for affective recognition tasks \cite{Eyben2015}.

Linguistic features have mainly been extracted through word-count-based programs like Linguistic Inquiry and Word Count (LIWC) \cite{Pennebaker2001} which is a software for extracting the count of words using an existing list of words and categories (e.g., positive/negative words, personal pronouns, social process). Its usage in couples research for example has shown the words partners utilize during conflict significantly affect interaction and overall marital quality. Findings indicate that greater first-person plural pronoun usage (‘we’), compared to first-person singular pronoun usage (‘I’) produces more positive resolutions to conflicts \cite{Simmons2005, Neysari2016}.  Tools such as LIWC however are not without their limitations. In fact, they depend on the accuracy and comprehensiveness of the dictionary they are based upon, together with not being able to take into account both the context that words are placed in and the different meanings they might hold \cite{Bantum2009}. In a context such as conflict interactions where specific word choices and their meanings are important in affecting how the conflict unfolds \cite{Simmons2005}, such limitations hold great significance for the validity and accuracy of applications that use such tools.  However, recent advances in natural language processing such as Bidirectional Encoder Representations and Transformations (BERT) \cite{Devlin2018} based on the Transformer architecture \cite{Vaswani2017} have been shown to set new state-of-the-art records in various natural language understanding tasks such as natural language inference, question answering, and sentiment analysis. Some prior works have evaluated the predictive capability of BERT relative to LIWC in psychotherapy and mental health classification with BERT  outperforming models based on LIWC features in populations with mental health diagnosis \cite{Tanana2021, Jiang2020}. Yet, BERT features have not yet been used in couples’ interaction research for prediction tasks. 

Some studies have used linguistic and or paralinguistic features specifically to predict behavioral codes for interacting romantic partners with the goal of automating behavioral coding. Most of these works have focused on session-level prediction — predicting one code for the whole 8-10 minute session \cite{black2010, lee2010, black2013, lee2014, xia2015, li2016, chakravarthula2015, tseng2016, tseng2017, chakravarthula2018, katsamanis2011, tseng2018} with a scarcity of works focused on prediction of fine-grained behavioral codes such as at the speaker turn level or every few seconds. One such work is that of Chakravarthula et al \cite{Chakravarthula2019} in which they trained machine learning models to predict 3 behavioral codes on a speaker turn level of 85 couples’ 10-minute conversations using paralinguistic features (from openSMILE) and linguistic features (custom sentence embedding model) and achieved 57.4\% unweighted average recall (balanced accuracy) for 3-class classification. Leveraging advanced sentence embedding methods such as BERT could potentially improve performance and increase the potential of automating behavioral coding. Yet, it has not been investigated in the context of couples research. Furthermore, including paralinguistic features could potentially enable better recognition.

In order to overcome current limitations, we utilized a data set collected from 368 couples (N = 736 participants) who were recorded during an 8-minute conflict interaction. Our main goal is to examine how linguistic and paralinguistic features in 10-second sequences can be used to predict how the same sequence is perceived and rated by human coders as positive or negative communication behavior. We aim to  answer the following research questions (RQs):

\textbf{RQ1:} \textit{Which linguistics features — LIWC or BERT — are better for predicting sequences-to-sequences-rated positive and negative communication behavior of partners?}

\textbf{RQ2:} \textit{Given that the raters focused on coding linguistic aspects of behavior, how does adding openSMILE’s eGeMAPS paralinguistic features affect the prediction performance?}

Our contributions are (1) an evaluation of the predictive capability of BERT vis-à-vis LIWC in the context of the automatic recognition of couples’ communication behavioral codes on a fine-grained time-scale (every 10 seconds) (2) an investigation into how the addition of paralinguistic features affects prediction performance (3) the use of a unique dataset — spontaneous, real-life, speech data collected from German-speaking Swiss couples (n=368 couples, N=736 participants), and the largest ever such dataset used in the literature for automatic coding of couples’ behavior. The insights from our work would enable the usage of new technologies to potentially automate the behavioral coding of couples, which could substantially improve the efficiency of couples research. 

\section{Methodology}

\textbf{Data Collection and Preprocessing:} This work used data from a larger dyadic interaction laboratory project conducted at the premises of the University of Zurich, Switzerland over 10 years with 368 heterosexual German-speaking, Swiss couples (N=736 participants; age 20-80) \cite{Kuster2015, uzh2020}. The inclusion criterion was to have been in the current relationship for at least 1 year. Couples had to choose one problematic topic for the conflict interaction from a list of common problems, and participants were then videotaped as they discussed the selected issue for 8 minutes. The data used in this work had one interaction from each couple and consequently, 368 8-minute interactions.

Two research assistants were trained to code communication behaviors using an adapted version of the Specific Affect Coding System (SPAFF) \cite{coan2007, Kuster2015}. Both raters practiced coding for at least 60 hours on videotapes that were not part of the study, with Cohen’s kappa indicating that they had achieved an acceptable interobserver agreement (k = 0.9). Each interaction was rated by both raters, with one rater focusing on the male partner and the other rater focusing on the female partner. Ratings were produced every 10 seconds to account for the behavior unfolding during each sequence, resulting in 48 sequences for each interaction. \textit{Positive communication}: (1) careful listening, interest, curiosity, (2) recognition, approval, factual praise, (3) affective communication, caring, (4) constructive criticism, and \textit{negative communication}:  (1) blaming, criticism, (2) defensiveness, (3) domineering, (4) withdrawal, stonewalling, (5) formally negative interaction, (6) contempt, (7) provocation, belligerence. For each 10-second sequence, raters would thus assign the code representing the communication behavior that was most prevalent out of the ones listed above. Raters were asked to focus on the verbal aspect of the behavior in assigning the codes. Due to the vast variety of codes present,  we categorized all types of positive and negative as 1 and 2 respectively, and then passed them to machine learning models in the form of a binary classification problem.

The speech was manually annotated with the start and end of each speaker’s turn, along with pauses and noise. The speech was manually transcribed in 15-second chunks separately for each partner. Given that Swiss German is mostly spoken with different dialects across Switzerland, the spoken words were written as the corresponding German word equivalent. Transcripts and audio recordings acquired from the interactions were divided along the same 10-second sequence to match the 10-second sequences used for behavioral coding. This process was done separately for each partner’s transcript and speech data. Consequently, we dropped 10-second matched transcript-audio-code sequences that contained no speech and transcribed words.

Of the original 368 Swiss heterosexual couples that took part in the study, we could only use 345 because some couples requested their data to be removed and some data were missing arising from technical problems in data collection. In addition, while the orignal dataset presented instances where behaviors had been coded as neutral/no communication, these were dropped from the analyses since no accurate description for what constituted neutral communication was given in the codebook, and no differentiation with instances of no communication was provided. This thus resulted in a total of 9930 10-seconds speech sequences with their matching behavioral codes. Out of that number, 6978 were instances where communication had been coded as positive, while 2952 were instances where it had been coded as negative, highlighting a significant class imbalance that is characteristic of real-world datasets and partners’ behavior as seen in other works (e.g., \cite{Chakravarthula2019}).

\textbf{Linguistic Features:} We extracted linguistic features from each 10-second transcript sequence using the LIWC software for German \cite{meier2019}. It utilizes an existing list of words and categories (e.g., positive/negative words, personal pronouns, social process) to count the number of the corresponding words in the transcript sequences and categorize them across 97 different features. The internal German LIWC dictionary was used to analyze the transcript and extract the features. We normalized each transcript sequence’s feature vector by dividing the value of all the other features by the “word count” feature which represents the number of words present in each transcript sequence. We then dropped the word count feature. This procedure thus left 96 normalized features that were passed as input to the machine learning models. 

Also, we extracted features from each 10-second sequence using a pretrained Sentence-BERT (SBERT) model \cite{Reimers2019}. Sentence-BERT is a modification of the BERT architecture with siamese and triplet networks to compute sentence embeddings such that semantically similar sentences are close in vector space. Sentence-BERT has been shown to outperform the mean and CLS token outputs of regular BERT models for semantic similarity and sentiment classification tasks. Given that the text is in German, we used the German BERT model \cite{germanbert} as SBERT’s Transformer model and the mean pooling setting. The German BERT model was pretrained using the German Wikipedia dump, the OpenLegalData dump, and German news articles. The extraction resulted in a  768-dimensional feature vector.

\textbf{Paralinguistic Features:} We extracted acoustic features from the voice recordings. For each 10-second sequence, we first used the speaker annotations to get the acoustic signal for each partner separately. Next, we used openSMILE \cite{Eyben2015} to extract the 88 eGeMAPS acoustic features which have been shown to be a minimalist set of features for affective recognition tasks  \cite{Eyben2010}. The original audio was encoded with 2 channels. As a result, we extracted the features for each channel resulting in a 176-dimensional feature vector. 

\section{Experiments and Evaluation}
We performed multiple experiments using the support vector machine (SVM) algorithm with the radial basis function (RBF) kernel and the scikit-learn library \cite{Pedregosa2011}. We used RBF since it performed the best in as our initial explorations in comparison to random forests, XGBoost, and linear SVM. We trained models to perform binary classification of the behavioral codes for positive and negative communication using different feature sets. Specifically, we used features from LIWC, BERT, openSMILE. Also, we explored multimodal fusion at the features level of BERT and openSMILE by concatenating features from both groups. We used TF-IDF unigram and bigram features (using the most frequent 1000 features) of the transcripts as a linguistic baseline. To train and evaluate the models, we used nested K-fold cross-validation (CV). The nested procedure consisted of utilizing an “inner” run of 3-fold CV for hyperparameter tuning, followed by an “outer” run of 5-fold CV which utilizes the best values for each hyperparameter found by the “inner” run. We prevented data from the same couple from being in both the train and test folds,  thereby evaluating the model’s performance on data from unseen couples. As the data was imbalanced, we utilized the metric balanced accuracy which is the unweighted average of the recall of each class, and confusion matrices for evaluation.  We used different values of the hyperparameter “C”, presenting results for the hyperparameter that produced the best results. We used the “balanced” hyperparameter for all the SVM models to mitigate the class imbalance while training. Standard errors were computed by running models repeatedly, randomizing the groups used for the K fold CV and thus gathering a set of 20 accuracy measures for each model.

\section{Results and Discussion}
Table \ref{tab:model_performances} presents the results of the best performing models for each of the feature modalities. The model that used only the BERT features performed the best with 69.4\% accuracy, compared to the LIWC model with 65.4\% accuracy. The BERT only approach also performed better than combining paralinguistic features to the BERT features. The latter performed closely with 69.2\% accuracy, but still significantly worse than the BERT only model (p<.001 on a Wilcoxon signed rank test). Indeed, the paralinguistic baseline approach using openSMILE features performed the worst, 61.3\%, being outperformed by TF-IDF linguistic baseline (65.6\%) which also slightly outperformed LIWC. The worse performance of the paralinguistic features is expected given that raters in the study were instructed to focus on the verbal aspect of the interaction rather than nonverbal behavior in assigning codes. 


Our results indicate that LIWC features do not have as much discriminative potential for prediction tasks compared both with even simpler approaches such as TF-IDF, as well as more advanced methods such as BERT. One likely explanation for BERT’s superior performance is its ability to capture the semantics of text via contextualized embeddings. These results have also been shown in a similar work using emotion psychotherapy or mental health data \cite{Tanana2021, Jiang2020}. The performance gain afforded by BERT notwithstanding, the simpler approaches did perform better than expected, with the performance improvement between TF-IDF and BERT being less than 4\%. It is worth noting that this BERT model is used out-of-the-box and it is out-of-domain without any customization on the couples conversational text. The results indicate that researchers in social psychology ought to consider alternatives to LIWC, such as BERT, for extracting features for prediction tasks such as automated behavioral coding and emotion recognition. Although LIWC features (and indeed, TF-IDF) have the advantage of being simpler and more easily interpretable, various approaches are being developed to make BERT features more interpretable via its multi-head attention mechanism \cite{Vig2019} and Shapley \cite{Kokalj2021}. Finally, the result of the BERT model performing better than the multimodal approach is consistent with other works that found a similar result for emotion \cite{Boateng2020b} and behavioral recognition \cite{Chakravarthula2019}. Including paralinguistic features did not seem to add any more predictive information especially considering the context of the study in which assigning codes focused on verbal behavior. Further approaches need to be explored to better combine the openSMILE and BERT features for improved results.

\begin{table}[]
\centering
\caption{Results across 20 runs (with standard errors) for models using LIWC, BERT, openSMILE and multimodal input features}
\label{tab:model_performances}
\begin{tabular}{|l|c|}
\hline
\textbf{Input Features} & \textbf{Balanced Accuracy (\% +/- S.E.)}  \\ \hline
openSMILE               & 61.28 $\pm$ .07                           \\ \hline
TF-IDF + ngrams         & 65.61 $\pm$ .08                           \\ \hline
LIWC                    & 65.41 $\pm$ .05                           \\ \hline
BERT                    & 69.39 $\pm$ .06                           \\ \hline
BERT and openSMILE      & 69.18 $\pm$ .06                           \\ \hline
\end{tabular}
\end{table}

  

\section{Limitations and Future Work}
In this work, we used manual transcripts. To accomplish true automated behavioral coding, our approach needs to use and work for automated transcriptions. Current speech recognition systems do not work for this unique dataset given that couples speak Swiss German, which is (1) a spoken dialect and not written, and (2) varies across different parts of the German-speaking regions of Switzerland. Further work is needed to develop automatic speech recognition systems for Swiss German.

Also, we only used the BERT model as a feature extractor to make a fair comparison with the LIWC features. Fine-tuning the BERT model on this task and domain to update the weights of the model would potentially improve the prediction results. This approach will be explored in future work. Finally, BERT models have been shown to encode gender and racial bias because of the data they are trained on. This consideration needs to be factored in for the specific prediction task and context \cite{bender2021}.

\section{Conclusion}
In this work, we investigated the predictive potential of BERT features for automated coding of couples’ communication behavior compared to LIWC features, the de facto linguistic features in social psychology. We extracted and compared LIWC and BERT features, used openSMILE features as a paralinguistic baseline and TF-IDF with ngrams as a linguistic baseline. We trained an RBF SVM to classify positive and negative communication behavior of each romantic partner on a 10-second granularity. Our results showed that both simple TF-IDF features as well as more complex BERT features both outperform LIWC, indicating that it might be time for researchers to consider alternatives to LIWC for predictive tasks in couples interactions. Additionally, adding paralinguistic features did not perform better than the BERT-only approach. Our work is a further step towards better approaches in automating the coding of couples’ behavior which could enhance couples research and assessments in couples therapy.

\begin{acks}
Funding was provided by the Swiss National Science Foundation: CR12I1\_166348/1; CRSI11\_133004/1; P3P3P1\_174466; P300P1\_164582
\end{acks}

\balance{}

\bibliographystyle{ACM-Reference-Format}
\bibliography{refs}

\end{document}